\documentclass{article}

\usepackage{microtype}
\usepackage{stfloats}
\usepackage{graphicx}
\usepackage{booktabs} % for professional tables

\usepackage{hyperref}

\usepackage[accepted]{icml2024}

\usepackage{amsmath}
\usepackage{amssymb}
\usepackage{mathtools}
\usepackage{amsthm}

\usepackage[capitalize,noabbrev]{cleveref}

\usepackage{multirow}
\usepackage{multicol}
\usepackage{caption}
\usepackage{subcaption}

\theoremstyle{plain}

\theoremstyle{definition}

\theoremstyle{remark}

\usepackage[textsize=tiny]{todonotes}

\icmltitlerunning{Training Foundation Models for Zero-Shot Multivariate Time Series Forecasting through Next Curve Shape Prediction}

\begin{document}

\twocolumn[
\icmltitle{Only the Curve Shape Matters: Training Foundation Models for Zero-Shot Multivariate Time Series Forecasting through Next Curve Shape Prediction}

% It is OKAY to include author information, even for blind
% submissions: the style file will automatically remove it for you
% unless you've provided the [accepted] option to the icml2024
% package.

% List of affiliations: The first argument should be a (short)
% identifier you will use later to specify author affiliations
% Academic affiliations should list Department, University, City, Region, Country
% Industry affiliations should list Company, City, Region, Country

% You can specify symbols, otherwise they are numbered in order.
% Ideally, you should not use this facility. Affiliations will be numbered
% in order of appearance and this is the preferred way.

\begin{icmlauthorlist}
\icmlauthor{Cheng Feng}{yyy}
\icmlauthor{Long Huang}{yyy,comp}
\icmlauthor{Denis Krompass}{yyy}
\end{icmlauthorlist}

\icmlaffiliation{yyy}{Siemens Technology}
\icmlaffiliation{comp}{Tsinghua University}
\icmlcorrespondingauthor{Cheng Feng}{cheng.feng@siemens.com}
\icmlcorrespondingauthor{Long Huang}{huangl22@mails.tsinghua.edu.cn}
\icmlcorrespondingauthor{Denis Krompass}{denis.krompass@siemens.com}
% You may provide any keywords that you
% find helpful for describing your paper; these are used to populate
% the "keywords" metadata in the PDF but will not be shown in the document
\icmlkeywords{Machine Learning, ICML}

\vskip 0.3in
]

% this must go after the closing bracket ] following \twocolumn[ ...

% This command actually creates the footnote in the first column
% listing the affiliations and the copyright notice.
% The command takes one argument, which is text to display at the start of the footnote.
% The \icmlEqualContribution command is standard text for equal contribution.
% Remove it (just {}) if you do not need this facility.

\printAffiliationsAndNotice{}  % leave blank if no need to mention equal contribution
% \printAffiliationsAndNotice{\icmlEqualContribution} % otherwise use the standard text.

\begin{abstract}
We present General Time Transformer (GTT), an encoder-only style foundation model for zero-shot multivariate time series forecasting. GTT is pretrained on a large dataset of 200M high-quality time series samples spanning diverse domains. In our proposed framework, the task of multivariate time series forecasting is formulated as a channel-wise next curve shape prediction problem, where each time series sample is represented as a sequence of non-overlapping curve shapes with a unified numerical magnitude. GTT is trained to predict the next curve shape based on a window of past curve shapes in a channel-wise manner. Experimental results demonstrate that GTT exhibits superior zero-shot multivariate forecasting capabilities on unseen time series datasets, even surpassing state-of-the-art supervised baselines. Additionally, we investigate the impact of varying GTT model parameters and training dataset scales, observing that the scaling law also holds in the context of zero-shot multivariate time series forecasting. 
\end{abstract}

\section{Introduction}
Time series forecasting, the task of predicting future values of one or multiple variables based on their historical values and other potentially relevant information, holds significant importance across diverse domains including manufacturing, traffic, healthcare, finance, and environmental science. In response to its practical significance, a large variety of time series forecasting methods have been developed. Earlier work includes classic statistical approaches such as ARIMA~\cite{box1968some,durbin2012time}, Exponential Smoothing~\cite{hyndman2008forecasting} and VAR~\cite{zivot2006vector}, as well as those leverage deep sequential models like recurrent neural networks (RNNs)~\cite{salinas2020deepar} and convolutional neural networks (CNNs) \cite{borovykh2018dilated}. In recent years, two distinct directions have emerged regarding the utilization of deep neural networks for time series forecasting. On one hand, building on the success of the Transformer architecture \cite{vaswani2017attention} in natural language processing (NLP) and computer vision (CV), there has been a surge in leveraging Transformer-like architecture for time series forecasting. Examples include Pyraformer~\cite{liu2021pyraformer}, LogTrans~\cite{li2019enhancing},  Informer~\cite{zhou2021informer}, Autoformer~\cite{wu2021autoformer}, FEDformer~\cite{zhou2022fedformer}, Crossformer~\cite{zhang2022crossformer} and PatchTST~\cite{nie2022time}, to name a few. On the other hand, there is also a different voice that simple multilayer perceptrons (MLP)-like models can achieve similar or even better time-series forecasting performance than sophisticated Transformer-based models~\cite{zeng2023Transformers,ekambaram2023tsmixer}. This discrepancy may be attributed to the fact that Transformers tend to overfit small datasets, and that the largest publicly available time series dataset is less than 10 GB~\cite{godahewa2021monash}, which is significantly smaller compared to those in NLP and CV domains.

In this work, inspired by the Transformer scaling successes in NLP and CV domains, we experiment with training a Transformer-based foundation model, which we term General Time Transformer (GTT), for zero-shot multivariate time series forecasting on a large dataset containing 200M high-quality time series samples collected from diverse domains. To overcome the challenges of dataset/distribution shift, as well as varying channel/variable\footnote{We will use the terms ``channel" and ``variable" interchangeably throughout the rest of the paper.} dimensions of time series samples across different domains, the task of multivariate time series forecasting is formulated as a channel-wise next curve shape prediction problem within our framework. Specifically, we do not introduce any time-series-specific inductive biases, but instead treat each time series sample as a sequence of non-overlapping curve shapes with a unified numerical magnitude. Each curve shape comprises M consecutive time points of a single variable. GTT is trained to use N preceding curve shapes as the context to predict the next curve shape on a channel-wise basis. We adopt an encoder-only architecture for GTT with the fewest possible modifications to the standard Transformer. The only major modification we introduced is a cross-channel attention stage after the temporal attention stage in each multi-head self-attention block to capture cross-variate dependency between channels. GTT employs an auto-regressive approach to handle long-term forecasting tasks extend beyond M time steps.

GTT exhibits excellent zero-shot multivariate time series forecasting performance on various benchmark datasets, even outperforming state-of-the-art supervised baselines in several cases.  We also demonstrate that GTT can achieve noticeably improved performance with cost-effective fine-tuning on target datasets. Moreover, in zero-shot univariate forecasting, GTT achieves higher prediction accuracy compared to other pretrained foundation models specifically designed for univariate time series forecasting. Additionally, we have conducted an investigation into the influence of different scales of GTT model parameters and training datasets, which reveals that the scaling law also holds in our context. Our codebase is available at https://github.com/cfeng783/GTT.

We summarize the main contributions of this paper as follows:
\begin{itemize}
    \item We introduce a framework that formulates the task of cross-domain multivariate time series forecasting into a problem of predicting the next curve shape given a context window of past curve shapes in a unified numerical magnitude on a channel-wise basis. This framework lays a solid foundation for the development of large-scale foundation models for multivariate time series forecasting.
    \item Our experimental results demonstrate that foundational models for time series forecasting, trained on datasets of comparable size to those used in CV and NLP domains, can also achieve outstanding zero-shot forecasting capabilities.
    \item To the best of our knowledge, GTT is the first foundation model for zero-shot \emph{multivariate} time series forecasting.
\end{itemize}

\section{Related Work}
% The main models for early time series forecasting are local univariant models based on statistics, such as ARIMA\cite{box1968some,durbin2012time} and Exponential Smoothing\cite{hyndman2008forecasting}, which fit each time series individually in a dataset. These models decompose time series into interpretable components (level/trend/seasonality) to make the prediction robust and understandble, but exhibit low efficiency when dealing with large-scale time series, and due to being trained on each time series independently, they fail to share similar patterns that may exist across different time series.
In recent years, deep neural networks such as RNNs~\cite{salinas2020deepar} and dilated CNNs~\cite{borovykh2017conditional} have gained significant prominence as formidable contenders in the field of time-series forecasting, particularly when confronted with large training datasets. Empirical evidence has demonstrated their superiority over conventional statistical methods, including ARIMA and exponential smoothing via various forecasting competitions \cite{makridakis2022m5,kopp2021traffic4cast}.

More recently, drawing inspiration from the triumph of the Transformer architecture in the realms of NLP and CV, there has been a notable surge in the utilization of Transformer-like architectures for time series forecasting. The first group of Transformer variants focuses on the development of novel attention modules to reduce computational complexity for long time series. Pyraformer~\cite{liu2021pyraformer}, LogTrans~\cite{li2019enhancing}, and Informer~\cite{zhou2021informer} are notable examples within this category. The second group of Transformer variants places specific emphasis on leveraging time and frequency domain features. For example, Autoformer~\cite{wu2021autoformer} introduces a seasonal-trend decomposition architecture, incorporating an auto-correlation mechanism to serve as an attention module. Recognizing that the point-wise attention of the Transformer architecture fails to capture overall characteristics of time series, FEDformer~\cite{zhou2022fedformer} proposes a frequency-enhanced seasonal-trend decomposition method to better capture global properties of time series. Three variants closely aligned with our research are Crossformer~~\cite{zhang2022crossformer}, iTransformer~\cite{liu2023itransformer} and PatchTST~\cite{nie2022time}. Crossformer and iTransformer, in particular, explicitly exploit cross-channel dependencies that play a vital role in achieving precise multivariate forecasting outcomes. PatchTST, on the other hand, employs patching techniques to enhance the local semantic information of input tokens of time series data.

Conversely, an alternative perspective suggests that simple MLP-like models may yield comparable, if not superior, performance in time-series forecasting when compared to sophisticated Transformer-based models~~\cite{zeng2023Transformers,ekambaram2023tsmixer}. Our conjecture is rooted in the notion that Transformers tend to overfit small datasets. For instance, the largest publicly available dataset for time series analysis is less than 10 GB~\cite{godahewa2021monash}, which pales in comparison to the vast datasets used in the NLP and CV domains to train Transformers. Thus, to better leverage the powerful modelling ability of Transformers while mitigating the risk of overfitting, reprogramming or fine-tuning pretrained acoustic and large language models (LLMs) for time series forecasting becomes another promising option \cite{yang2021voice2series,zhou2023one,jin2023time,chang2023llm4ts}. Meanwhile, there are also works which directly use LLMs for time series forecasting \cite{gruver2023large}.

The exploration of foundation models pretrained on large datasets for zero-shot time series forecasting remains relatively limited in comparison to the advancements made in NLP and CV fields. However, there have been some notable efforts recently. Among the examples are ForecastPFN~\cite{dooley2023forecastpfn}, TimeGPT~\cite{garza2023timegpt}, Lag-Llama~\cite{rasul2023lag} and PreDcT~\cite{das2023decoder}. ForecastPFN is a Transformer-based prior-data fitted network trained purely on synthetic data designed to mimic common time series patterns. TimeGPT is a Transformer-based time series forecasting model trained over 100B data points, with other data and model details remain unrevealed. Lag-Llama, a probabilistic time series forecast model adapted from the LlaMA~\cite{touvron2023llama} architecture, is trained on a large collection of time series from the Monash Time Series Repository \cite{godahewa2021monash}. PreDcT is a patched-decoder style model trained on 1B time points from Google Trends. GTT distinguishes itself from existing models through several notable differences. Firstly, our training data is much more diverse compared with ForecastPFN, LlaMA and PreDcT. Secondly, we utilize an encoder-only architecture instead of a decoder-only architecture, wherein the task of time series forecasting is approached as a problem of predicting the next curve shape in a unified numerical magnitude. Lastly, GTT incorporates a channel attention mechanism, specifically designed for multivariate time series forecasting, rather than focusing solely on univariate forecasting.

% TimeGPT is trained on the largest publicly available collection of time series data, utilizing a Transformer architecture. On the other hand, the work referenced as [21] focuses on pretraining a patched-decoder style attention model. Lag-Llama, on the other hand, constructs lagged features to train a foundation model based on the LlaMA architecture (Touvron et al., 2023). These models have demonstrated remarkable accuracy in forecasting across diverse domains and applications, requiring no additional training. Moreover, they offer enhanced accessibility, accuracy, and efficiency, significantly reducing computational complexity and time requirements.

% Foundation models pre-trained on a large datasets remains relatively under-explored for time series forecasting tasks compared with in NLP and CV. But there are some parallel works focused on foundation model for time series recently like TimeGPT\cite{garza2023timegpt}, decoder-only foundation model\cite{das2023decoder} and Lag-Llama\cite{rasul2023lag}. TimeGPT is trained in the largest collection of publicly available time series based on Transformer architecture while [21] is based on pretraining a patched-decoder style attention model. Lag-Llama constructs lagged features to train a foundation model based on the LlaMA\cite{touvron2023llama} architecture. These model produce accurate predictions across a diverse array of domains and applications without additional training and are more accessible and accurate, less time-consuming, and drastically reduces computational complexity.

\section{Problem Definition}
We consider building a general purpose zero-shot multivariate time series forecaster that takes in a look-back window of $L$ time points of a time-series and optionally their corresponding time features as context, and predicts the future $H$ time points. Let $\mathbf{x}_{1:L}$ and $\mathbf{d}_{1:L}$ be the context time series and corresponding time feature values, GTT is a function to predict $\hat{\mathbf{x}}_{L+1:L+H}$, such that 
\begin{equation*}
\hat{\mathbf{x}}_{L+1:L+H} = f(\mathbf{x}_{1:L},\mathbf{d}_{1:L})
\end{equation*}
Note since we are building a general purpose multivariate forecaster, the only covariates we consider in the pretraining stage are three time features: second of the day, day of the week and month of the year. These three time features, if available, are converted to 6 features using sine and cosine transformations\footnote{https://ianlondon.github.io/blog/encoding-cyclical-features-24hour-time/}.

\section{Method}

\subsection{Pretraining Data Preparation}
\label{sec:data}
We collected a large scale time series repository containing 2.4B univariate or multivariate time points from both internal and public sources. Our repository consists of about 180,000 univariate or multivariate time series spanning diverse domains including manufacturing, transportation, finance, environmental sensing, healthcare, to name some.

For each series, we take its first $90\%$ time points to extract training samples and the remaining $10\%$ time points to extract validation samples (we monitor validation loss for early stopping during training). Each extracted time series sample consists of 1088 consecutive time points without missing values. Our model is trained to predict the values of the last 64 time points using the preceding 1024 time points as context.

To ensure consistency, we restrict the max number of channels for a time series sample to 32, in which 6 channels are reserved for time features. In case the number of channels of a time series sample is less than 32, we complement its channel number to 32 by setting all the values in the added channels to zero. Conversely, if a time series sample has more than 32 channels, we divide it into samples with 32 or fewer channels and then supplement the samples with less than 32 channels to reach the total of 32 channels. 

% Furthermore, we pad 960 time points with zero values in the beginning of each dataset to generate samples with shorter context lengths. To distinguish between originally exist zero values and manually added zero values, we add a small value ($10^{-5}$) to all originally exist values.

To achieve a unified numerical magnitude for time series samples across different datasets, we normalize each time series sample on a channel-wise basis. Specifically, the first 1024 time points are normalized to have zero mean and unit variance. Then, the last 64 time points are normalized using the calculated channel mean and standard deviation from the first 1024 time points. More precisely, let $x_{1:1024}$ and $x_{1025:1088}$ be the first 1024 and last 64 time points for a channel of time series sample, the normalization is conducted as follows:
\begin{eqnarray*}
x_{1025:1088} &=& \frac{x_{1025:1088} - \text{mean}(x_{1:1024})}{\text{stdev}(x_{1:1024})+\epsilon} \\  
x_{1:1024} &=& \frac{x_{1:1024} - \text{mean}(x_{1:1024})}{\text{stdev}(x_{1:1024})+\epsilon}
\end{eqnarray*} Normalized samples that have a data point with an absolute value greater than 9 are discarded to exclude samples containing extreme values. Furthermore, we mask 1 to 960 time points in the beginning of $10\%$ randomly chosen samples to zero values. This manipulation allows us to generate samples with shorter context lengths, providing a variation in the length of context within the training data.

Lastly, to ensure a balance between the scale and domain diversity of our training data, we restrict the max number of training or validation samples that can be extracted from a single time series to 60,000. In the end, approximately 200M high quality training samples and 24M validation samples are generated from our repository.

\begin{figure*}[t]
\centering
\centerline{\includegraphics[width=\linewidth]{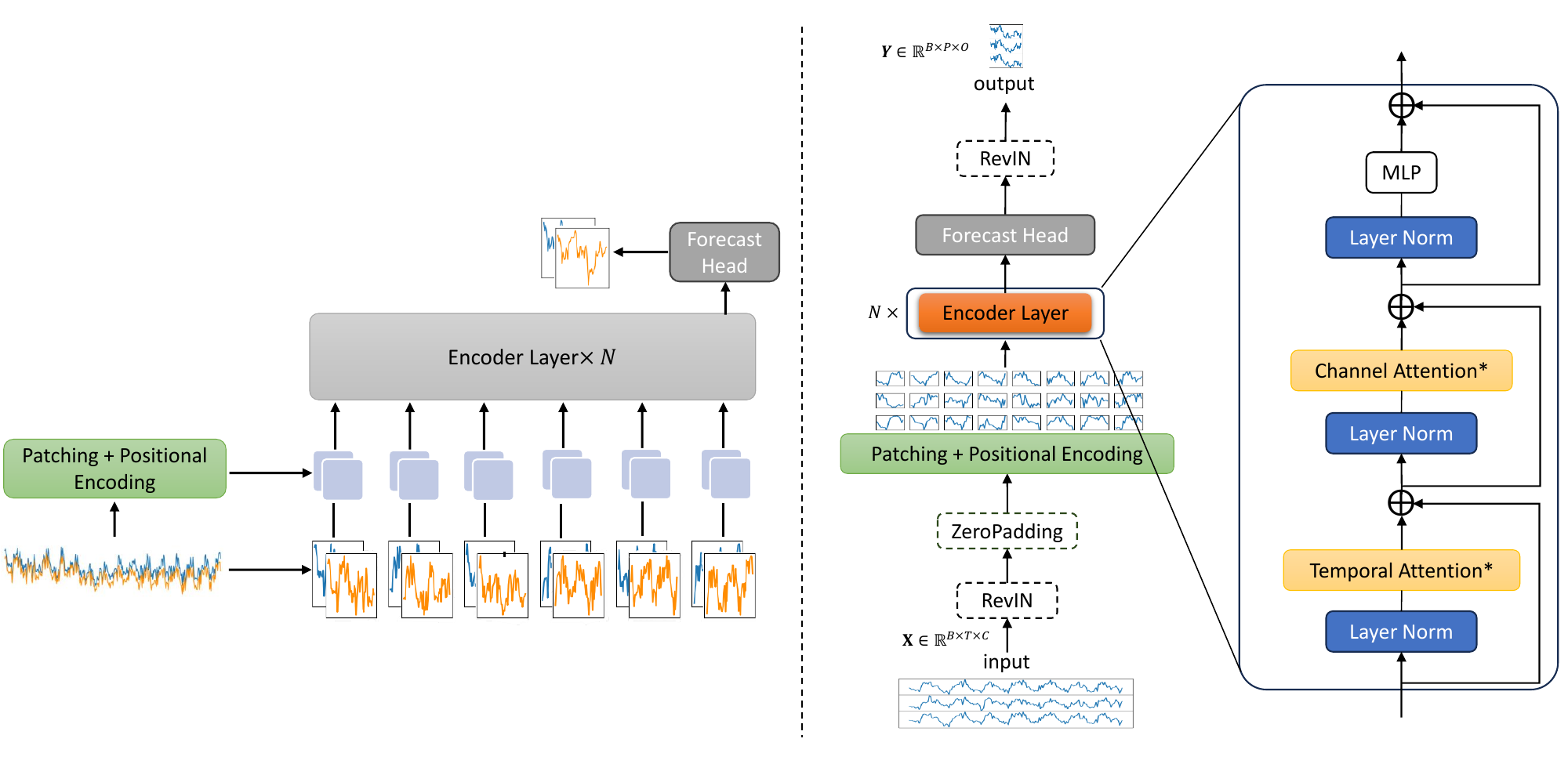}}
\caption{Model overview. We split an input multivariate time series into fixed-size non-overlapping patches (curve shapes) channel-wise, linearly embed each of them, add position encodings, and feed the resulting sequence of patches to the encoder. The encoder has an extra channel attention stage compared with the standard Transformer, the temporal and channel attention share the same weights. We add a linear head to the last token to perform forecasting. During inference, we add RevIN layers to normalize and denormalize time series channels and pad zeros in front of time series samples with less than 1024 time points.}
\label{fig:model}
\end{figure*}

\subsection{General Time Transformer}
An overview of GTT is depicted in Figure~\ref{fig:model}. Specifically, we split an input multivariate time series sample into fixed-size non-overlapping patches channel-wise. Each patch represents a curve shape composed of 64 time points of a single variable. We linearly embed each of the patches, add position encodings, and then feed the resulting sequence of patches to the encoder. The encoder has an extra channel attention stage compared with the standard Transformer. For parameter efficiency, the temporal and channel attention share the same weights. Weight sharing between multi-head self attentions applied on different input dimensions has been proved effective~\cite{yang2022aim}. Lastly, we add a linear head to the last token to perform forecasting of the next patch (curve shape). During inference, we apply reversible instance normalization (RevIN)~\cite{kim2021reversible} to first normalize the input data into zero mean and unit variance and then padding zeros in the front if the length of input time series is shorter than 1024. The predicted output is denormalized into its original scale by using the pre-calculated mean and variance of the input data. In this way, no normalization is needed for input data before using GTT which significantly improves the convenience of model usage. 

It is important to note that upon closer examination, the architectural similarities between GTT and Vision Transformer (ViT) \cite{dosovitskiy2020image} become apparent if curve shapes are thought as special type of image patches. However, a significant distinction arises in their treatment of channels. While ViT combines RGB channels of an image within its patching process, GTT independently processes time series channels and incorporates an additional stage for channel attention, which facilitates the learning of cross-variate dependencies with varying channel numbers. We now describe the key components of GTT architecture. In the presentation, we use the following notations: $B$: batch size, $T$: input time series length, $C$: number of input channels (number of target variables, covariates, time features in total), $O$: number of output channels (number of target variables), $M$: number of patches, $P$: patch size, $D$: number of embedding dimensions, $N$: number of encoder layers.

\paragraph{Patching and Positional Encoding} Let $\mathbf{X} \in \mathbb{R}^{B\times T \times C}$ be the input batch of time series samples, we first reshape $\mathbf{X}$ to $\hat{\mathbf{X}} \in \mathbb{R}^{BC \times T \times 1}$, then utilize a one-dimensional convolutional layer (Conv1D) with kernel size and strides equal to patch size $P$ and number of filters equals to $D$, to segment input series into patches and then embed them into $M \times D$ dimensional patch embeddings channel-wise. We use the Positional Encoding in the original Transformer paper~\cite{vaswani2017attention} for encoding position information and add position encodings to the patch embeddings to retain sequential information:
\begin{small}
\begin{align*}
&\hat{\mathbf{X}}= \text{Reshape}(\mathbf{X}),  \quad \mathbf{X} \in \mathbb{R}^{B\times T \times C},  \hat{\mathbf{X}}\in \mathbb{R}^{BC \times T \times 1}    \\
&\mathbf{Z}_0= \text{Conv1D}(\hat{\mathbf{X}}) + \mathbf{E}_{pos}, \quad   \mathbf{E}_{pos}, \mathbf{Z}_0 \in \mathbb{R}^{BC \times M \times D}
\end{align*}
\end{small}

\begin{figure*}[t]
\centering
\begin{subfigure}{.5\textwidth}
  \centering
  \includegraphics[width=.9\linewidth]{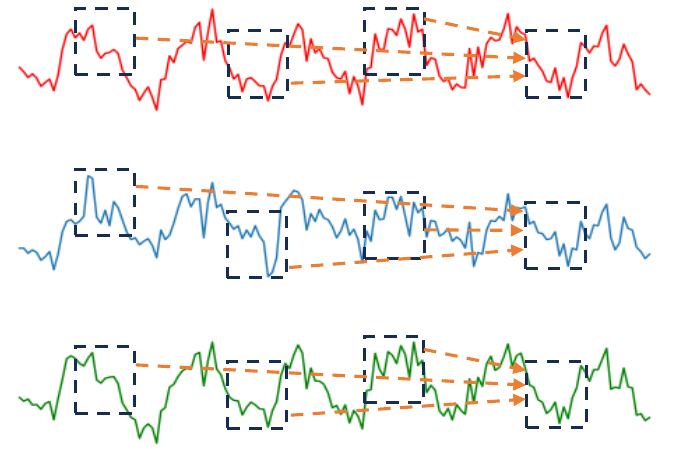}
  \caption{Temporal Attention (T-MSA)}
  \label{fig:sub1}
\end{subfigure}%
\begin{subfigure}{.5\textwidth}
  \centering
  \includegraphics[width=.9\linewidth]{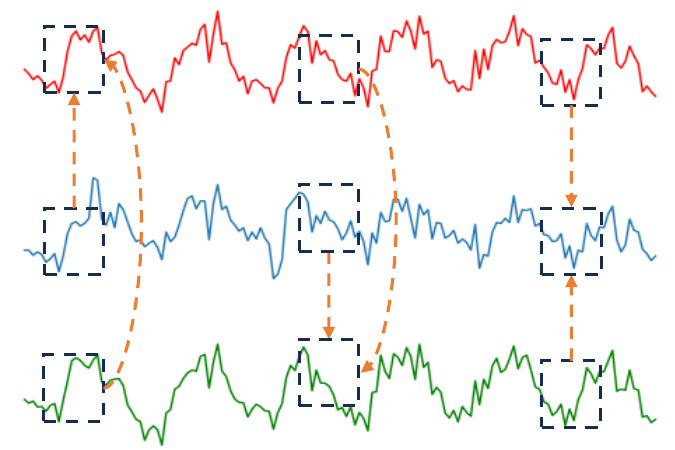}
  \caption{Channel Attention (C-MSA)}
  \label{fig:sub2}
\end{subfigure}
\caption{Illustration of Temporal and Channel Attention}
\label{fig:MSA}
\end{figure*}

\paragraph{Encoder Layers}
To utilize both temporal and cross channel dependencies for the forecasting task, we apply two multi-head self attentions (MSA), which we call temporal attention (T-MSA) and channel attention (C-MSA) in each encoder layer of GTT. An illustration of T-MSA and C-MSA is given in Figure~\ref{fig:MSA}. A MLP consisting of two layers with a GELU non-linearity \cite{hendrycks2016gaussian} is applied after T-MSA and C-MSA. Layernorm (LN) and residual connections are also applied:
% \begin{small}
\begin{align*}
&\mathbf{Z}'_l = \text{T-MSA}( \text{LN}(\mathbf{Z}_{l-1}) ) + \mathbf{Z}_{l-1}, \quad l=1,\ldots,N \\
&\hat{\mathbf{Z}}'_l = \text{Reshape}(\mathbf{Z}'_l), \quad \mathbf{Z}'_l \in  \mathbb{R}^{BC \times M \times D}, \hat{\mathbf{Z}}'_l \in \mathbb{R}^{BM \times C \times D}\\
&\hat{\mathbf{Z}}''_l = \text{C-MSA}( \text{LN}(\hat{\mathbf{Z}}'_l) ) + \hat{\mathbf{Z}}'_l, \quad l=1,\ldots,N \\
&\mathbf{Z}''_l = \text{Reshape}(\hat{\mathbf{Z}}''_l), \quad \hat{\mathbf{Z}}''_l \in  \mathbb{R}^{BM \times C \times D}, \mathbf{Z}''_l \in \mathbb{R}^{BC \times M \times D}\\
&\mathbf{Z}_l = \text{MLP}( \text{LN}(\mathbf{Z}''_{l}) )+\mathbf{Z}''_l , \quad l=1,\ldots,N
\end{align*}
% \end{small}
Note that although the channel number in the pretraining stage is set to 32, since channel attention requires no positional information, the trained model can generalize to varying channel dimensions in the inference stage.

\paragraph{Forecast Head}
Following the encoder layers, we retrieve $\mathbf{Z}_N^M$, the last token of the last encoder layer, and then a linear forecast head is attached to $\mathbf{Z}_N^M$ for predicting the next patch of time points for all channels, i.e., the linear head is shared by all channels. Lastly, we retrieve the channels for the target variables as the output:
\begin{align*}
&\mathbf{Y}' = \mathbf{Z}_N^M  W^{D \times P} + \mathbf{b}^P, \quad \mathbf{Z}_N^M \in  \mathbb{R}^{BC \times  D}, \mathbf{Y}' \in \mathbb{R}^{BC \times P} \\
&\mathbf{Y}'' = \text{Reshape}(\mathbf{Y}'), \quad \mathbf{Y}'' \in \mathbb{R}^{B \times P \times C}\\
&\mathbf{Y} = \text{Retrieve}(\mathbf{Y}''), \quad \mathbf{Y} \in \mathbb{R}^{B \times P \times O}\\
\end{align*}
% \end{small}
\paragraph{Loss Function}
The model is trained to minimize the Mean Absolute Error (MAE) between ground-truth and predicted values. We choose the MAE loss because it is less sensitive to outliers. It is worth to mention that the MAE loss is only calculated on the originally exist data points, i.e., data points in the supplemented channels from the data preparation step are excluded from the loss computation.
% nonzero part of the ground-truth values. That is to say, we set error to zero wherever the ground-truth value is zero. This is because these zero values are manually added in the data preparation step (we added a small value ($10^{-5}$) to all originally exist values to distinguish between originally exist zero values and manually added zero values in the data preparation step), thus should not be counted in back-propagation.

\paragraph{RevIN and Zero-Padding}
During the inference phase, we employ reversible instance normalization (RevIN)~\cite{kim2021reversible} to normalize the input data to have zero mean and unit variance within the model. If the length of the input time series is shorter than 1024, we pad zeros at the beginning. The predicted output is then denormalized back to its original scale using the pre-calculated mean and variance of the input data. This approach eliminates the need for normalization of the input data prior to using GTT, thereby significantly improving the convenience of model usage during inference.

It is important to note that RevIN is not employed during the training stage, which distinguishes it from many recently proposed deep supervised models~\cite{wu2022timesnet,nie2022time,zhou2023one,chang2023llm4ts}. This decision is based on the fact that the use of RevIN cannot guarantee a unified magnitude of values for time series samples. For instance, two time series samples [0.0001,0.0002,...,0.1288] and [0.001,0.002,...,1.288], with the same curve shapes, would result in the exact same loss value in our framework. However, if RevIN were used, they would contribute different loss values if their original values were used to calculate the loss. This discrepancy would introduce bias towards time series samples with larger values during pretraining.

\subsection{Why Encoder-only Architecture}
Our encoder-only architecture ensures that the predicted values are normalized strictly using the mean and standard deviation of the entire context window. However, employing a decoder-only architecture is problematic in this regard. In the decoder-only architecture, where the first patch predicts the second patch, and subsequently, the first two patches predicts the third patch, the normalization process faces a conflict. Specifically, during the pretraining data preparation step, normalization is based on the mean and standard deviation of the complete context window. As a result, the normalization of predicted values for earlier patches is influenced by the mean and standard deviation of subsequent patches which are not observable during the inference phase if the decoder-only architecture were used. This conflict compromises the normalization process, leading to inconsistencies in the magnitude of values across different time series samples.

% the second patch is based on the mean and standard deviation of all patches in the context window during training. However, during the inference phase, the subsequent patches are not observed.

% Specifically, when using the decoder architecture where the first patch predicts the second patch, and subsequently, the second patch predicts the third patch, the predicted values for the second patch are normalized based on the mean and standard deviation of all patches in the context window during training, whereas the later patches are unobservable during the inference phase. 

\begin{table*}
    \centering
    \caption{Details of GTT model variants.}
    \begin{tabular}{cccccc}
         \toprule
         Model & Encoder layers & Embedding dimension & Number of heads & MLP size & Parameters \\
         \midrule
         GTT-Tiny & 4 & 384 & 6 & 1536 & 7M  \\
         GTT-Small & 6 & 512 & 8 & 2048 & 19M  \\
         GTT-Large & 8 & 768 & 12 & 3072 & 57M  \\
         \bottomrule
    \end{tabular}
    \label{tab:variants}
\end{table*}

\begin{table*}
    \centering
    \caption{Statistics of benchmark datasets.}
    \begin{tabular}{ccccc}
         \toprule
         Datasets  & Number of features & Time points (Train, Validation, Test) & Frequency   \\
         \midrule
         ETTm1, ETTm2 & 7 & (34465, 11521, 11521) & 15 mins  \\
         ETTh1, ETTh2 & 7 & (8545, 2881, 2881) & Hourly  \\
         Electricity & 321 & (18317, 2633, 5261) & Hourly  \\
         Traffic & 862 & (12185, 1757, 3509) & Hourly  \\
         Weather & 21 & (36792, 5271, 10540) & 10 mins  \\
         ILI & 7 & (617, 74, 170) & Weekly  \\
         \bottomrule
    \end{tabular}
    \label{tab:data}
\end{table*}

\section{Experiments}
\subsection{Experimental Settings}
\paragraph{Model Variants}
Our largest trained model has 57M parameters, which is significantly smaller than those foundation models in NLP and CV domains. Nevertheless, we already observe excellent zero-shot forecasting performance. Details on the GTT model variants are provided in Table~\ref{tab:variants}. All models are trained using the 200M training samples and 24M validation samples as described in Section~\ref{sec:data}. We train all models using the AdamW optimizer~\cite{loshchilov2017decoupled}, training is stopped when the validation loss increases in three consecutive epochs. More training details are given in the appendix.

\paragraph{Benchmark Datasets} To evaluate the forecasting performance of GTT, we follow the benchmarks used in PatchTST~\cite{nie2022time}. Specifically, 8 popular datasets, including 4 ETT datasets, Electricity, Traffic, Weather and ILI are used. The statistics of benchmark datasets are summarized in Table~\ref{tab:data}. It is also worthy to mention that all the benchmark datasets are not included in our pretraining data.

% We divide the ETT dataset into train/val/test by the ratio of 12:4:4 months. The train/val/test for Electricity, Traffic, Weather is 0.7:0.1:0.2 and for Exchange, ILI is 0.7:0.1:1.

\begin{table*}
    \begin{center}
    \caption{Multivariate time series forecasting. The results are obtained by averaging predictions for four different lengths: 24, 36, 48, and 60 for the ILI dataset, 96, 192, 336, and 720 for other datasets. The best value for each metric is highlighted in red, while the second-best value is highlighted in blue. ZS is short for Zero-Shot and FT is short for Fine-Tune.}
    \resizebox{\textwidth}{!}{
    \label{tab:mulret}
    \tabcolsep=0.0065\linewidth
    \begin{tabular}{c|cc|cc|cc|cc|cc|cc|cc|cc|cc|cc}
    \toprule
    \multirow{2}{*}{Model} & \multicolumn{2}{c|}{GTT (ZS)}& \multicolumn{2}{c|}{GTT (FT)} & \multicolumn{2}{c|}{GPT4TS} & \multicolumn{2}{c|}{PatchTST} & \multicolumn{2}{c|}{Crossformer} & \multicolumn{2}{c|}{Fedformer} & \multicolumn{2}{c|}{TimesNet} &\multicolumn{2}{c|}{DLinear}   & \multicolumn{2}{c|}{TSMixer} & \multicolumn{2}{c}{iTransformer}\\ 
    & MSE & MAE & MSE & MAE & MSE & MAE & MSE & MAE & MSE & MAE & MSE & MAE & MSE & MAE & MSE & MAE & MSE & MAE & MSE & MAE\\
    \midrule
    ETTm1 & 0.398 & 0.392 & 0.370 & 0.383 &\textcolor{blue}{0.352} &0.383 &0.353 &0.382 &0.405 &0.424 &0.448 &0.452 &0.400 &0.406 &0.357 &\textcolor{blue}{0.379} &\textcolor{red}{0.351} &\textcolor{red}{0.378}&- &-\\
    ETTm2 & 0.279 & 0.324 & \textcolor{red}{0.253} & \textcolor{red}{0.309} &0.266 &0.326 &0.256 &0.317 &- &- &0.305 &0.349 &0.291 &0.333 &0.267 &0.332 &\textcolor{blue}{0.254} &\textcolor{blue}{0.314} &- &-\\
    ETTh1 & 0.418& \textcolor{blue}{0.415} & 0.420 & \textcolor{red}{0.411} &0.427 &0.426 &\textcolor{blue}{0.413} &0.434 &0.457 &0.454 &0.440 &0.460 &0.458 &0.450 &0.423 &0.437 &\textcolor{red}{0.408} &0.430&- &-\\
    ETTh2 &\textcolor{blue}{0.314} &\textcolor{blue}{0.359} &\textcolor{red}{0.298}& \textcolor{red}{0.353} &0.346 &0.394 &0.331 &0.381 &- &- &0.434 &0.447 &0.414 &0.427 &0.431 &0.447 &0.340 &0.387&- &-\\
    Electricity & \textcolor{blue}{0.157} &\textcolor{blue}{0.249} & \textcolor{red}{0.155} & \textcolor{red}{0.246} &0.167 &0.263 &0.159 &0.253 &0.305 &0.358 &0.214 &0.327 &0.192 &0.295 &0.166 &0.264 &\textcolor{red}{0.155} &0.251 &0.178 &0.270\\
    Traffic &0.404 &\textcolor{blue}{0.260} & \textcolor{blue}{0.390} & \textcolor{red}{0.257} &0.414 &0.294 &0.391 &0.264 &0.506 &0.285 &0.610 &0.376 &0.620 &0.336 &0.434 &0.295 &\textcolor{red}{0.386} & 0.263&0.428 &0.282\\
    Weather &0.227 &\textcolor{red}{0.247} & \textcolor{red}{0.218} & \textcolor{blue}{0.249} &0.237 &0.270 &0.226 &0.264 &0.409 &0.447 &0.309 &0.360 &0.259 &0.287 &0.246 &0.300 &\textcolor{blue}{0.224} &0.262&0.258 &0.279\\
    % Exchange &0.599 &0.498 &- &- &- &- &- &- &0.519 &0.500 &\textcolor{blue}{0.416} &\textcolor{blue}{0.443} &\textcolor{red}{0.354} &\textcolor{red}{0.414} &- &-\\
    ILI &\textcolor{blue}{1.536} & \textcolor{blue}{0.732} & 1.668 & \textcolor{red}{0.724} &1.925 &0.903 &\textcolor{red}{1.480} &0.807 &3.387 &1.236 &2.847 &1.144 &2.139 &0.931 &2.169 &1.041 &- &- &- &-\\
    \bottomrule
    \end{tabular}
    }
    \end{center}
\end{table*}

\subsection{Comparison to Supervised Models}
We first compare the zero-shot multivariate forecasting performance of our largest model, GTT-Large, to state-of-the-art supervised forecasting models, including GPT4TS~\cite{zhou2023one}, PatchTST~\cite{nie2022time}, Crossformer~\cite{zhang2022crossformer}, Fedformer~\cite{zhou2022fedformer}, TimesNet~\cite{wu2022timesnet}, DLinear~\cite{zeng2023Transformers}, TSMixer~\cite{ekambaram2023tsmixer} and iTransformer~\cite{liu2023itransformer}. All the above supervised baselines are trained on the train split of each benchmark dataset. Additionally, we also report the performance of GTT (we refer GTT to GTT-Large if not specified hereafter) after fine-tuning on the train split of each benchmark dataset. Note that we only tune the Forecast Head and keep other parameters of GTT fixed during fine-tuning. The results, in terms of Mean Square Error (MSE) and Mean Absolute Error (MAE) on the test split of each benchmark dataset, are given in Table~\ref{tab:mulret}. We report the results of the baselines directly from their original papers if available.

We find that GTT without fine-tuning achieves the best MAE on 1 dataset, second best MAE on 5 datasets, and second best MSE on 2 datasets. Notably, GTT performs remarkably well even in a zero-shot scenario, where it faces a disadvantage as other methods have the opportunity to train on the benchmark datasets. Furthermore, we find that after fine-tuning on the train split of benchmark datasets, the performance of GTT can be further significantly improved. It achieves the best MAE on 6 datasets, best MSE on 4 datasets. These results clearly demonstrate the superiority of GTT as a foundation model for multivariate time series forecasting.

\begin{table}
    \begin{center}
    \caption{Zero-shot univariate time series forecasting for the ``OT" feature of benchmark datasets. The results are obtained by averaging predictions for four different lengths: 24, 36, 48, and 60 for the ILI dataset, 96, 192, 336, and 720 for other datasets. The best number for each metric is colored red.}
    % \resizebox{\textwidth}{!}{
    \label{tab:uni1}
    % \tabcolsep=0.0065\linewidth
    \begin{tabular}{c|cc|cc}
    \toprule
    \multirow{2}{*}{Model} & \multicolumn{2}{c|}{GTT} & \multicolumn{2}{c}{ForecastPFN}\\
    & MSE & MAE & MSE & MAE\\
    \midrule
    ETTm1 & \textcolor{red}{0.049} & \textcolor{red}{0.160} &0.175 &0.322\\
    ETTm2 & \textcolor{red}{0.118} & \textcolor{red}{0.254} &0.568 &0.598\\
    ETTh1 & \textcolor{red}{0.084} & \textcolor{red}{0.222} &0.190 &0.350\\
    ETTh2 & \textcolor{red}{0.212} & \textcolor{red}{0.359} &0.604 &0.618\\
    Electricity &\textcolor{red}{0.267} &\textcolor{red}{0.355} &2.257 &1.233\\
    Traffic &\textcolor{red}{0.126} &\textcolor{red}{0.219} &4.121 &1.650\\
    Weather &\textcolor{red}{0.002} &\textcolor{red}{0.028} &0.003 &0.040\\
    % Exchange &0.526 &0.515 &0.535 &0.405 &- &- &\textcolor{red}{0.058} &-\\
    ILI &\textcolor{red}{0.582} &\textcolor{red}{0.551} &2.903 &9.825\\
    \bottomrule
    \end{tabular}
    % }
    \end{center}
\end{table}

\begin{table}
    \begin{center}
    \caption{Zero-shot univariate time series forecasting for the ``OT" feature of benchmark datasets. The prediction length is 96 for ETT datasets, and 24 for ILI. The best number for each metric is colored red. The results for PreDcT is cited from~\cite{das2023decoder}.}
    % \resizebox{\textwidth}{!}{
    \label{tab:uni2}
    % \tabcolsep=0.0065\linewidth
    \begin{tabular}{c|cc|cc}
    \toprule
    \multirow{2}{*}{Model} & \multicolumn{2}{c|}{GTT} & \multicolumn{2}{c}{PreDcT}\\
    & NRMSE & WAPE & NRMSE & WAPE\\
    \midrule
    ETTm1 & \textcolor{red}{0.129} & \textcolor{red}{0.096} &0.591 &0.310 \\
    ETTm2 & 0.240 & 0.174 &\textcolor{red}{0.199} &\textcolor{red}{0.123} \\
    ETTh1 & \textcolor{red}{0.203} & \textcolor{red}{0.153} &0.672 &0.378 \\
    ETTh2 & 0.397 & 0.306 &\textcolor{red}{0.238} &\textcolor{red}{0.151} \\
    ILI &\textcolor{red}{0.313} & 0.215 &0.477 &\textcolor{red}{0.152} \\
    \bottomrule
    \end{tabular}
    % }
    \end{center}
\end{table}

% \begin{figure}
% \centering
% \includegraphics[width=.95\linewidth]{scale_mae.eps}
% \label{fig:scale1}
% \caption{Zero-shot multivariate forecasting performance on benchmark datasets of GTT with different model parameter scales. The results for ETT are averaged from the four ETT datasets.}
% \includegraphics[width=.95\linewidth]{dataset_mae.eps}
% \label{fig:scale2}
% \caption{Zero-shot multivariate forecasting performance on benchmark datasets of GTT-Large with different training data scales. The results for ETT are averaged from the four ETT datasets.}
% \end{figure}

\begin{figure}
\centering
  \includegraphics[width=.9\linewidth]{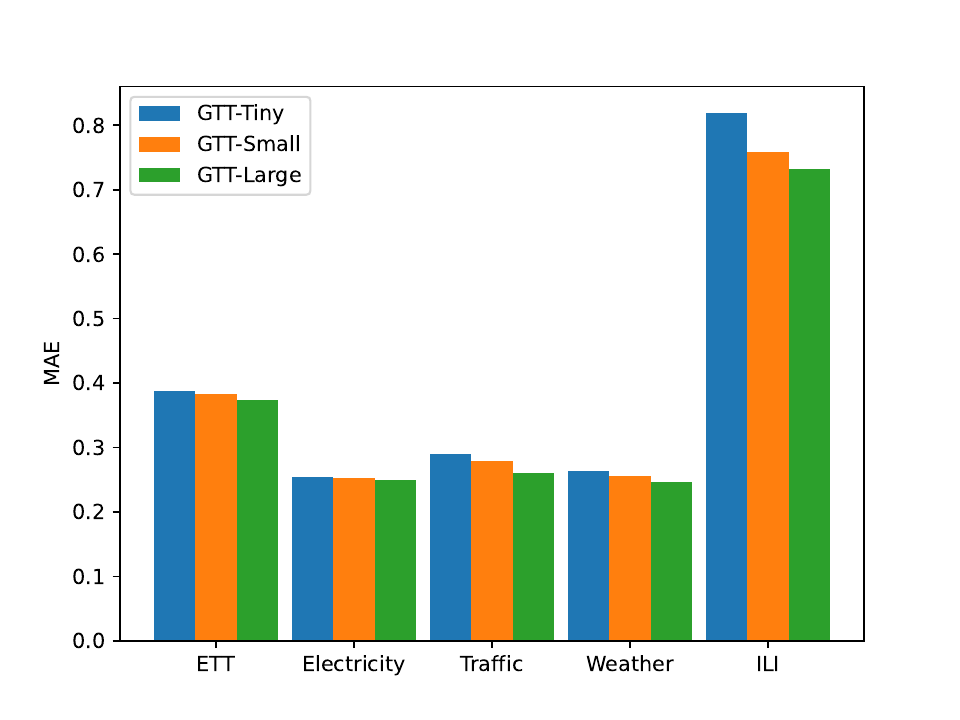}
\caption{Zero-shot multivariate forecasting performance on benchmark datasets of GTT with different model parameter scales. The results for ETT are averaged from the four ETT datasets.}
\label{fig:scale1}
\end{figure}

\begin{figure}
\centering
  \includegraphics[width=.9\linewidth]{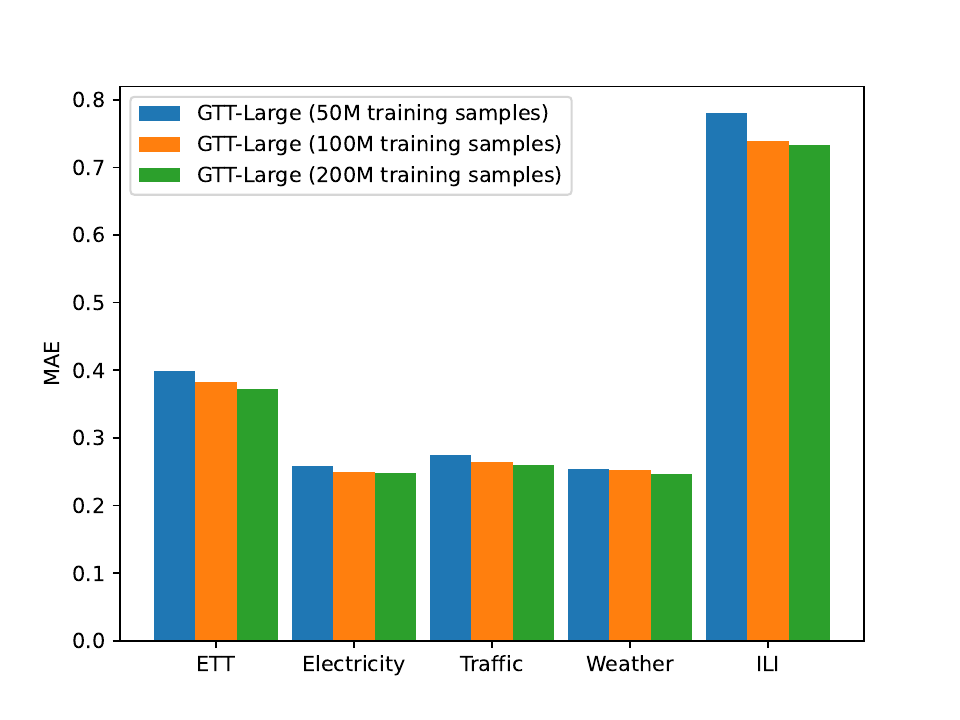}
\caption{Zero-shot multivariate forecasting performance on benchmark datasets of GTT-Large with different training data scales. The results for ETT are averaged from the four ETT datasets.}
\label{fig:scale2}
\end{figure}

\subsection{Comparison to Pretrained Models}
We then compare GTT against recently proposed pretrained models for zero-shot univariate time series forecasting, specifically ForecastPFN~\cite{dooley2023forecastpfn} and PreDcT~\cite{das2023decoder}. It should be noted that we excluded TimeGPT~\cite{garza2023timegpt} and Lag-Llama~\cite{rasul2023lag} from the comparison as their models have not been released yet and no comparable results on the benchmark datasets are presented in their papers. Table~\ref{tab:uni1} presents the zero-shot performance of GTT and ForecastPFN in terms of MAE and MSE on all benchmark datasets for univariate time series forecasting. GTT demonstrates a substantial margin of improvement over ForecastPFN, outperforming it on all datasets. Table~\ref{tab:uni2} compares GTT's zero-shot performance for univariate time series forecasting with PreDcT in terms of Normalized Root MSE (NRMSE) and Weighted Average Percentage Error (WAPE) on five benchmark datasets. It is important to note that since a pretrained model for PreDcT has not been released, we directly cite the results for PreDcT from the original paper~\cite{das2023decoder}. GTT outperforms PreDcT in half of the scenarios. It is worth mentioning that the comparison is limited to univariate forecasting as GTT is the only pretrained model capable of supporting both univariate and multivariate forecasting.

% \begin{figure}
% \centering
% \begin{subfigure}{.48\textwidth}
%   \centering
%   \includegraphics[width=.9\linewidth]{scale_mae.eps}
%   \caption{models trained on 120M training datasets (MAE)}
%   \label{fig:sub4}
% \end{subfigure}
% \begin{subfigure}{.48\textwidth}
%   \centering
%   \includegraphics[width=.9\linewidth]{dataset_mae.eps}
%   \caption{small model trained on different sizes datasets (MAE)}
%   \label{fig:sub4}
% \end{subfigure}
% \caption{Illustration of models of different scales trained on 120M training datasets}
% \label{fig:fig3}
% \end{figure}

% \begin{figure*}
% \centering
% \begin{subfigure}{.5\textwidth}
%   \centering
%   \includegraphics[width=.9\linewidth]{dataset_mse.eps}
%   \caption{small model trained on different sizes datasets (MSE)}
%   \label{fig:sub3}
% \end{subfigure}%
% \begin{subfigure}{.5\textwidth}
%   \centering
%   \includegraphics[width=.9\linewidth]{dataset_mae.eps}
%   \caption{small model trained on different sizes datasets (MAE)}
%   \label{fig:sub4}
% \end{subfigure}
% \caption{Illustration of small model trained on different sizes training datasets}
% \label{fig:fig4}
% \end{figure*}

\begin{figure*}
\centering
\begin{subfigure}{.32\textwidth}
  \centering
  \includegraphics[width=.9\linewidth]{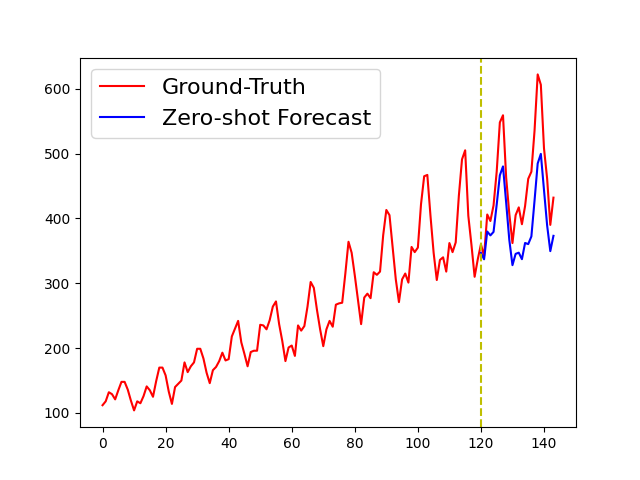}
\end{subfigure}%
\begin{subfigure}{.32\textwidth}
  \centering
  \includegraphics[width=.9\linewidth]{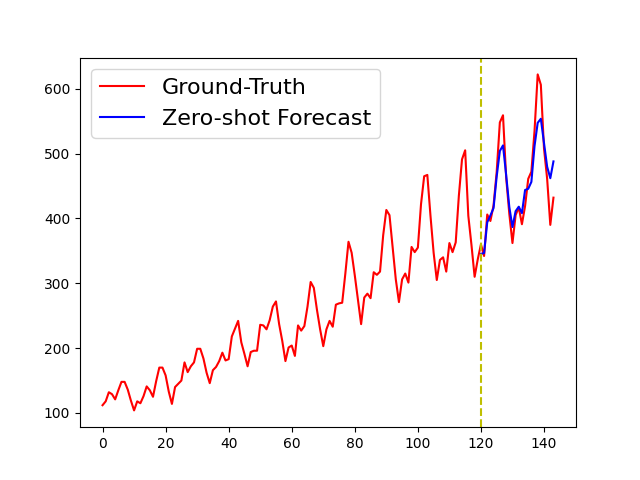}
\end{subfigure}
\begin{subfigure}{.32\textwidth}
  \centering
  \includegraphics[width=.9\linewidth]{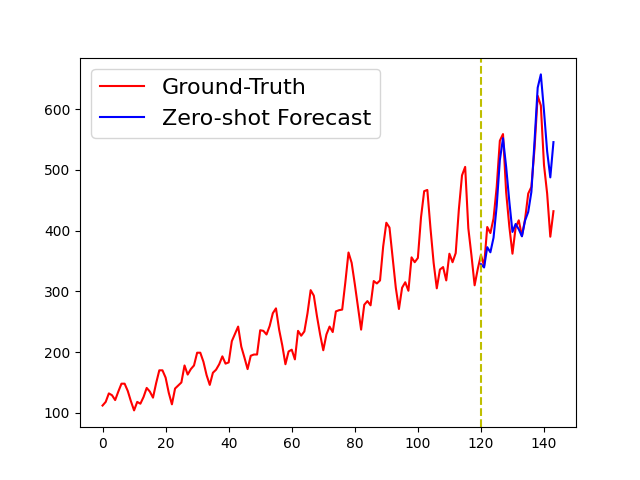}
\end{subfigure}
% \begin{subfigure}{.24\textwidth}
%   \centering
%   \includegraphics[width=.99\linewidth]{AP_large2.png}
% \end{subfigure}
\caption{Zero-shot forecast of last 24 months' values in Air Passenger dataset produced by GTT-Tiny (left), GTT-Small (mid), and GTT-Large (right). We observe that with a larger model, the accelerated increasing trend can be better captured.}
% \caption{Zero-shot forecast of last 24 months' values in Air Passenger dataset produced by GTT-Tiny (left), GTT-Small (mid left), GTT-Large (mid right) and GTT-Large without using time related features (right). We observe that 1) with a larger model, the accelerated increasing trend can be better captured. 2) Forecasting accuracy is significantly compromised if time related features are discarded.}
\label{fig:air}
\end{figure*}

\subsection{Scaling Study}
We first study how the parameter scale impacts the zero-shot multivariate forecast performance of GTT models. Figure~\ref{fig:scale1} gives the forecasting performance in terms of MAE on the benchmark datasets for GTT-Tiny, GTT-Small and GTT-Large. It can be observed that when the training data size does not bottleneck, the zero-shot forecasting accuracy increases with a larger model. Figure~\ref{fig:air} gives an interesting example of how model parameter scale impact zero-shot forecasting performance using the Air Passenger dataset \footnote{https://www.kaggle.com/datasets/rakannimer/air-passengers/data}: we find that with a larger model, the accelerated increasing trend in the Air Passenger dataset can be better captured.

We then study how crucial is the training dataset size. Specifically, we also pre-train GTT-Large models on smaller datasets of size: 50M and 100M training samples. Figure~\ref{fig:scale2} gives the zero-shot forecasting performance in terms of MAE on the benchmark datasets for GTT-Large pretrained on 50M, 100M and 200M samples respectively. It can be seen that when the model size does not bottleneck, the zero-shot forecasting accuracy also increases with a larger training dataset. 

Importantly, the above results indicates that GTT does not appear to reach saturation within the range of model parameter and dataset sizes explored. This motivates future scaling efforts for our proposed method.

\section{Conclusions}
We have explored training a Transformer-based foundation model called GTT for multivariate time series forecasting. We do not introduce any time-series-specific inductive biases into the GTT architecture. Instead, we interpret an arbitrary multivariate time series as a sequence of curve shapes (patches) and process it on a channel-wise basis by a Transformer encoder with an extra channel attention stage. 
This straightforward yet scalable approach yields impressive results, particularly when combined with pretraining on large datasets. GTT demonstrates comparable or superior performance to many advanced supervised models in multivariate time series forecasting across various widely used benchmark datasets.

While these initial findings are promising, several challenges still need to be addressed. Firstly, it is crucial to incorporate uncertainty calibration, encompassing both aleatoric and epistemic uncertainties, into the forecasting outcomes to enhance the reliability of GTT predictions. Another challenge is to extend the context length of GTT models to further improve their forecasting capabilities. Lastly, further scaling of GTT is expected to yield enhanced performance.

% Additionally, it is essential to explore the application of GTT to other time series analysis tasks, such as time series classification and anomaly detection.

\section{Potential Broader Impact}
This paper presents work whose goal is to advance the field of time series forecasting. One of the societal consequences of our work could be the promotion of large-scale foundation models specifically designed for time series analysis tasks. Like in CV and NLP domains, the reliability of such foundation models must be carefully checked before massive usage in public.

\bibliography{example_paper}
\bibliographystyle{icml2024}

%%%%%%%%%%%%%%%%%%%%%%%%%%%%%%%%%%%%%%%%%%%%%%%%%%%%%%%%%%%%%%%%%%%%%%%%%%%%%%%
%%%%%%%%%%%%%%%%%%%%%%%%%%%%%%%%%%%%%%%%%%%%%%%%%%%%%%%%%%%%%%%%%%%%%%%%%%%%%%%
% APPENDIX
%%%%%%%%%%%%%%%%%%%%%%%%%%%%%%%%%%%%%%%%%%%%%%%%%%%%%%%%%%%%%%%%%%%%%%%%%%%%%%%
%%%%%%%%%%%%%%%%%%%%%%%%%%%%%%%%%%%%%%%%%%%%%%%%%%%%%%%%%%%%%%%%%%%%%%%%%%%%%%%
\newpage
\appendix
\onecolumn

\section{Experiment Details}
Table~\ref{tab:pretraining} summarizes our pretraining setups for GTT variants. All GTT variants are trained on a cluster of Nvidia A800 GPUs. Pretraining is stopped when the validation loss increases in three consecutive epochs. 

For fine-tuning, we use the Adam optimizer \cite{kingma2014adam} with $10^{-3}$ learning rate (LR) and no LR decay. Fine-tuning is also stopped when the validation loss increases in three consecutive epochs. The model with the best validation loss is used for further experiments.
  
\begin{table}
    \centering
    \begin{tabular}{cccccccc}
         \toprule
         Models & Optimizer & Initial LR & LR Decay & Weight Decay & Gradient Clip Norm & Warmup Steps & Batch Size \\
         \midrule
         GTT-Large & AdamW & $3 \times 10^{-4}$ & cosine & 0.004 & 1.0 & 2048 & 1024 \\
         GTT-Small & AdamW & $6 \times 10^{-4}$ & cosine & 0.004 & 1.0 & 2048 & 2048 \\
         GTT-Tiny & AdamW & $1 \times 10^{-3}$ & cosine & 0.004 & 1.0 & 2048 & 4096 \\
         \bottomrule
    \end{tabular}
    \caption{Pretraining Setups for GTT models.}
    \label{tab:pretraining}
\end{table}

\section{Detailed Experiment Results}
We give the detailed results for multivariate time series forecasting on the benchmark datasets in Table \ref{tab:multifull}. The detailed results for univariate time series forecasting on benchmark datasets is presented in Table~\ref{tab:unifull}.
\begin{table*}
    \begin{center}
    \caption{Detailed results for multivariate time series forecasting. We fix the input context length of GTT to 128 for ILI, and 1024 for other datasets.}
    \resizebox{1.0\textwidth}{!}{
    \label{tab:multifull}
    % \tabcolsep=0.0055\linewidth
    \begin{tabular}{cc|cc|cc|cc|cc|cc|cc}
    \toprule
    \multicolumn{2}{c|}{\multirow{3}{*}{Model}} & \multicolumn{2}{c|}{GTT-Tiny} & \multicolumn{2}{c|}{GTT-Small} & \multicolumn{2}{c|}{GTT-Large} & \multicolumn{2}{c|}{GTT-Large} & \multicolumn{2}{c|}{GTT-Large} & \multicolumn{2}{c}{GTT-Large} \\
    & &\multicolumn{2}{c|}{} &\multicolumn{2}{c|}{} &\multicolumn{2}{c|}{} &\multicolumn{2}{c|}{(100M traing samples)} &\multicolumn{2}{c|}{(50M traing samples)} &\multicolumn{2}{c}{(Fine-tune)}\\
    & & MSE & MAE & MSE & MAE & MSE & MAE & MSE & MAE & MSE  & MAE & MSE & MAE\\
    \midrule
    \multirow{5}{*}{\rotatebox{90}{ETTm1}} 
    &96 &0.348 &0.362 &0.329 &0.353 &0.324 &0.352 &0.342  &0.366 &0.387  &0.394 &0.322 &0.355\\
    &192 &0.395 &0.389 &0.377 &0.384 &0.374 &0.381 &0.393 &0.398 &0.441  &0.428 &0.350  &0.373\\
    &336 &0.442 &0.413 &0.424 &0.412 &0.418 &0.403 &0.437  &0.424 &0.456  &0.486 &0.382  &0.391\\
    &720 &0.506 &0.443 &0.503 &0.456 &0.478 &0.432 &0.500  &0.459 &0.554  &0.496 &0.429  &0.417\\
    &mean &0.423 &0.402 &0.408 &0.401 &0.398 &0.392 &0.418  &0.412 &0.467  &0.444 &0.370  &0.383\\
    \midrule
    \multirow{5}{*}{\rotatebox{90}{ETTm2}}
    &96 &0.203 &0.270 &0.200 &0.269 &0.178 &0.256 &0.200  &0.268 &0.196  &0.271 &0.176  &0.256\\
    &192 &0.273 &0.316 &0.270 &0.315 &0.247 &0.304 &0.261  &0.309 &0.267  &0.319 &0.225 &0.290\\
    &336 &0.331 &0.353 &0.328 &0.352 &0.307 &0.344 &0.316  &0.344 &0.332  &0.362 &0.272  &0.323\\
    &720 &0.400 &0.398 &0.412 &0.406 &0.383 &0.395 &0.395  &0.394 &0.421  &0.419 &0.340  &0.370\\
    &mean &0.302 &0.334 &0.303 &0.336 &0.279 &0.324 &0.293  &0.329 &0.304  &0.343 &0.253  &0.309\\
    \midrule
    \multirow{5}{*}{\rotatebox{90}{ETTh1}}
    &96 &0.396 &0.393 &0.391 &0.390 &0.375 &0.384 &0.380  &0.380 &0.391  &0.391 &0.366  &0.377\\
    &192 &0.444 &0.420 &0.437 &0.413 &0.414 &0.407 &0.436  &0.409 &0.444  &0.420 &0.410  &0.401\\
    &336 &0.466 &0.436 &0.459 &0.427 &0.424 &0.419 &0.468  &0.432 &0.475  &0.444 &0.433  &0.418\\
    &720 &0.522 &0.476 &0.506 &0.461 &0.460 &0.450 &0.524  &0.475 &0.548  &0.496 &0.474  &0.451\\
    &mean &0.457 &0.431 &0.448 &0.423 &0.418 &0.415 &0.452  &0.424 &0.465  &0.438 &0.420 &0.411\\
    \midrule
    \multirow{5}{*}{\rotatebox{90}{ETTh2}}
    &96 &0.274 &0.328 &0.268 &0.317 &0.251 &0.310 &0.262  &0.312 &0.275  &0.319 &0.236  &0.308\\
    &192 &0.321 &0.362 &0.316 &0.352 &0.295 &0.342 &0.315  &0.349 &0.320  &0.354 &0.275  &0.335\\
    &336 &0.356 &0.390 &0.343 &0.378 &0.318 &0.366 &0.350  &0.379 &0.383  &0.355 &0.309  &0.362\\
    &720 &0.433 &0.445 &0.426 &0.435 &0.393 &0.421 &0.419  &0.433 &0.419  &0.432 &0.374  &0.411\\
    &mean &0.346 &0.381 &0.338 &0.371 &0.314 &0.359 &0.336  &0.368 &0.342  &0.372 &0.298  &0.353\\
    \midrule
    \multirow{5}{*}{\rotatebox{90}{Electricity}}
    &96 &0.119 &0.217 &0.123 &0.217 &0.117 &0.212 &0.119  &0.214 &0.124  &0.219 &0.116  &0.212\\
    &192 &0.140 &0.236 &0.147 &0.240 &0.138 &0.232 &0.140  &0.233 &0.148  &0.239 &0.136  &0.230\\
    &336 &0.164 &0.258 &0.170 &0.260 &0.162 &0.253 &0.162  &0.253 &0.174  &0.263 &0.159 &0.251\\
    &720 &0.219 &0.303 &0.211 &0.292 &0.215 &0.298 &0.213  &0.295 &0.238  &0.312 &0.212  &0.293\\
    &mean &0.161 &0.254 &0.163 &0.252 &0.157 &0.249 &0.159  &0.249 &0.171  &0.258 &0.155  &0.246\\
    \midrule
    \multirow{5}{*}{\rotatebox{90}{Traffic}}
    &96 &0.388 &0.257 &0.366 &0.247 &0.358 &0.235 &0.366  &0.241 &0.369  &0.246 &0.346  &0.232\\
    &192 &0.413 &0.271 &0.389 &0.260 &0.381 &0.246 &0.385  &0.251 &0.391  &0.260 &0.368  &0.244\\
    &336 &0.450 &0.292 &0.420 &0.281 &0.407 &0.262 &0.409  &0.266 &0.416  &0.277 &0.393  &0.259\\
    &720 &0.534 &0.338 &0.502 &0.329 &0.472 &0.298 &0.470  &0.299 &0.482  &0.318 &0.455  &0.295\\
    &mean &0.446 &0.289 &0.419 &0.279 &0.404 &0.260 &0.407  &0.264 &0.415  &0.275 &0.390  &0.257\\
    \midrule
    \multirow{5}{*}{\rotatebox{90}{Weather}}
    &96 &0.147 &0.183 &0.149 &0.182 &0.144 &0.179 &0.150  &0.183 &0.150  &0.183 &0.143  &0.182\\
    &192 &0.199 &0.235 &0.198 &0.231 &0.190 &0.224 &0.199  &0.231 &0.200  &0.231 &0.187  &0.228\\
    &336 &0.265 &0.284 &0.257 &0.277 &0.247 &0.267 &0.259  &0.274 &0.258  &0.272 &0.236  &0.269\\
    &720 &0.376 &0.352 &0.344 &0.334 &0.326 &0.319 &0.337  &0.323 &0.350  &0.330 &0.308 &0.319\\
    &mean &0.246 &0.264 &0.237 &0.256 &0.227 &0.247 &0.236  &0.253 &0.240  &0.254 &0.218  &0.249\\
    \midrule
    \multirow{5}{*}{\rotatebox{90}{ILI}}
    &24 &1.596 &0.733 &1.583 &0.718 &1.377 &0.676 &1.585  &0.686 &1.541  &0.701 &1.580  &0.677\\
    &36 &1.908 &0.810 &1.607 &0.740 &1.469 &0.714 &1.643  &0.720 &1.752  &0.763 &1.617  &0.707\\
    &48 &1.950 &0.835 &1.630 &0.760 &1.543 &0.740 &1.715  &0.753 &1.854  &0.798 &1.671  &0.735\\
    &60 &2.154 &0.900 &1.841 &0.818 &1.758 &0.800 &1.850  &0.799 &2.082  &0.862 &1.804  &0.780\\
    &mean &1.902 &0.819 &1.666 &0.759 &1.536 &0.732 &1.698  &0.739 &1.807  &0.781 &1.668  &0.724\\
    \bottomrule
    \end{tabular}
    }
    \end{center}
\end{table*}

\begin{table}
    \begin{center}
    \caption{Detailed results for univariate time series forecasting. We fix the input context length of GTT to 128 for ILI, and 1024 for other datasets.}
    % \resizebox{1.0\textwidth}{!}{
    \label{tab:unifull}
    % \tabcolsep=0.0055\linewidth
    \begin{tabular}{cc|cccc|cc}
    \toprule
    \multicolumn{2}{c|}{\multirow{2}{*}{Model}} & \multicolumn{4}{c|}{GTT-Large} & \multicolumn{2}{c}{ForecastPFN} \\ 
    & & MSE & MAE & NRMSE & WAPE& MSE & MAE\\
    \midrule
    \multirow{5}{*}{\rotatebox{90}{ETTm1}} 
    &96 &0.029 &0.127 &0.129 &0.096 &0.171 &0.317\\
    &192 &0.041 &0.149 &0.153 &0.112 &0.174 &0.321\\
    &336 &0.053 &0.167 &0.173 &0.126 &0.177 &0.325\\
    &720 &0.071 &0.196 &0.198 &0.146 &0.176 &0.324\\
    &mean &0.049 &0.160 &0.163 &0.120 &0.175 &0.322\\
    \midrule
    \multirow{5}{*}{\rotatebox{90}{ETTm2}}
    &96 &0.069 &0.190 &0.240 &0.174 &0.565 &0.599\\
    &192 &0.098 &0.232 &0.285 &0.211 &0.568 &0.599\\
    &336 &0.128 &0.269 &0.325 &0.244 &0.570 &0.598\\
    &720 &0.177 &0.326 &0.378 &0.293 &0.568 &0.596\\
    &mean &0.118 &0.254 &0.307 &0.231 &0.568 &0.598\\
    \midrule
    \multirow{5}{*}{\rotatebox{90}{ETTh1}}
    &96 &0.063 &0.189 &0.203 &0.153 &0.198 &0.358\\
    &192 &0.075 &0.211 &0.219 &0.169 &0.184 &0.345\\
    &336 &0.087 &0.229 &0.230 &0.179 &0.186 &0.347\\
    &720 &0.114 &0.262 &0.249 &0.194 &0.191 &0.350\\
    &mean &0.084 &0.222 &0.225 &0.174 &0.190 &0.350\\
    \midrule
    \multirow{5}{*}{\rotatebox{90}{ETTh2}}
    &96 &0.137 &0.285 &0.397 &0.306 &0.608 &0.619\\
    &192 &0.175 &0.327 &0.440 &0.345 &0.566 &0.598\\
    &336 &0.219 &0.374 &0.471 &0.376 &0.588 &0.612\\
    &720 &0.318 &0.450 &0.512 &0.408 &0.655 &0.642\\
    &mean &0.212 &0.359 &0.455 &0.359 &0.604 &0.618\\
    \midrule
    \multirow{5}{*}{\rotatebox{90}{Electricity}}
    &96 &0.217 &0.314 &0.562 &0.378 &2.206 &1.221\\
    &192 &0.249 &0.336 &0.602 &0.405 &2.165 &1.206\\
    &336 &0.281 &0.363 &0.641 &0.438 &2.251 &1.231\\
    &720 &0.322 &0.408 &0.694 &0.500 &2.405 &1.272\\
    &mean &0.267 &0.355 &0.625 &0.430 &2.257 &1.233\\
    \midrule
    \multirow{5}{*}{\rotatebox{90}{Traffic}}
    &96 &0.103 &0.186 &0.269 &0.156 &4.173 &1.654\\
    &192 &0.110 &0.198 &0.278 &0.166 &4.090 &1.646\\
    &336 &0.124 &0.219 &0.294 &0.183 &4.111 &1.649\\
    &720 &0.171 &0.275 &0.342 &0.228  &4.111 &1.651\\
    &mean &0.126 &0.219 &0.296 &0.183 &4.121 &1.650\\
    \midrule
    \multirow{5}{*}{\rotatebox{90}{Weather}}
    &96 &0.001 &0.022 &0.445 &0.317 &0.002 &0.038\\
    &192 &0.001 &0.027 &0.520 &0.383 &0.002 &0.038\\
    &336 &0.002 &0.030 &0.574 &0.431 &0.003 &0.040\\
    &720 &0.002 &0.036 &0.683 &0.513 &0.003 &0.043\\
    &mean &0.002 &0.028 &0.556 &0.411 &0.003 &0.040\\
    \midrule
    \multirow{5}{*}{\rotatebox{90}{ILI}}
    &24 &0.471 &0.470 &0.313 &0.215 &1.102 &0.887\\
    &36 &0.522 &0.520 &0.321 &0.231 &1.071 &0.893\\
    &48 &0.605 &0.572 &0.334 &0.245 &1.210 &0.949\\
    &60 &0.732 &0.640 &0.354 &0.265 &1.447 &1.032\\
    &mean &0.582 &0.551 &0.331 &0.239 &1.208 &0.940\\
    \bottomrule
    \end{tabular}
    % }
    \end{center}
\end{table}

\section{Visualization of Zero-shot Forecasting Results}
\begin{figure*}
\centering
\begin{subfigure}{.37\textwidth}
  \centering
  \includegraphics[width=\linewidth]{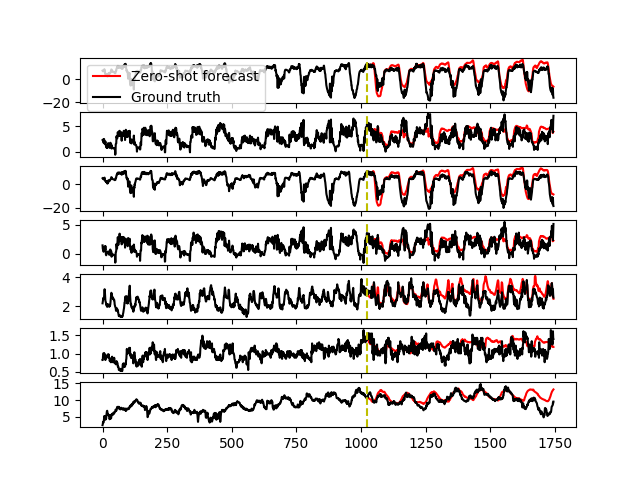}
  \caption{ETTm1}
  \label{fig:sub3}
\end{subfigure}%
\begin{subfigure}{.37\textwidth}
  \centering
  \includegraphics[width=\linewidth]{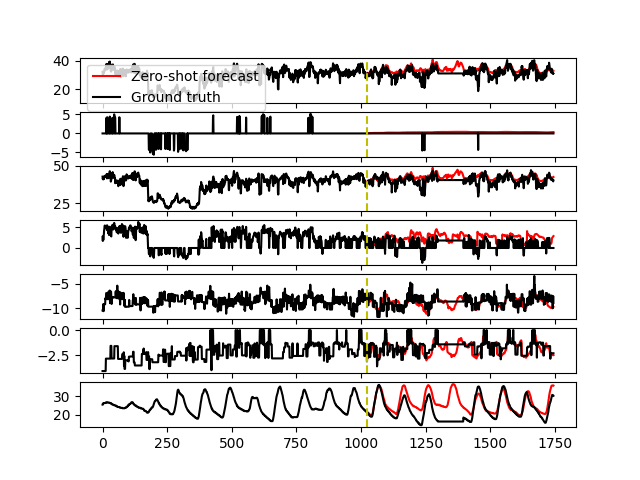}
  \caption{ETTm2}
  \label{fig:sub3}
\end{subfigure}%
\\
\begin{subfigure}{.37\textwidth}
  \centering
  \includegraphics[width=\linewidth]{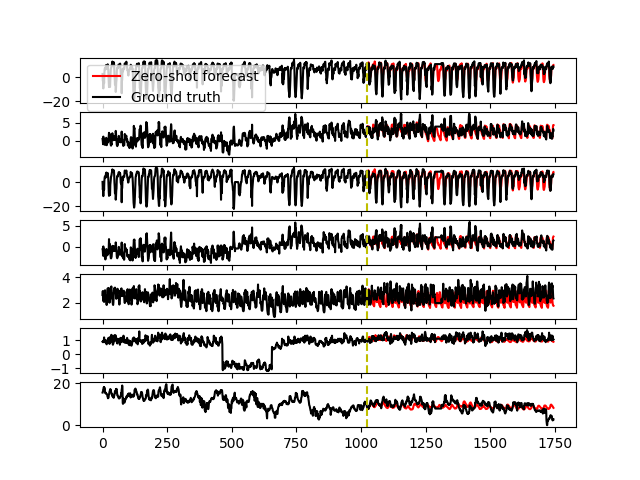}
  \caption{ETTh1}
  \label{fig:sub3}
\end{subfigure}%
\begin{subfigure}{.37\textwidth}
  \centering
  \includegraphics[width=\linewidth]{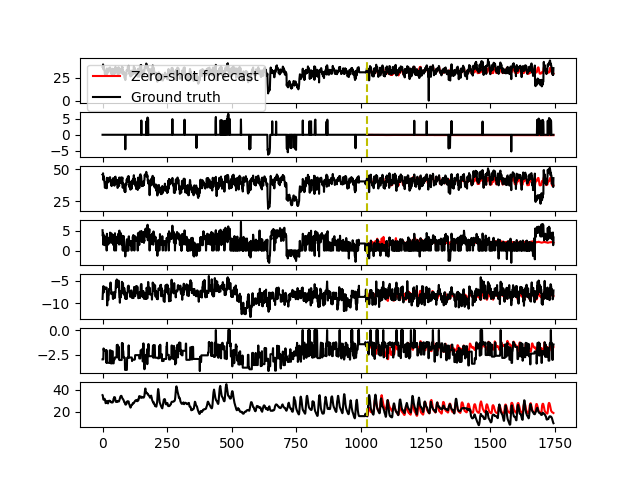}
  \caption{ETTh2}
  \label{fig:sub3}
\end{subfigure}%
\\
\begin{subfigure}{.37\textwidth}
  \centering
  \includegraphics[width=\linewidth]{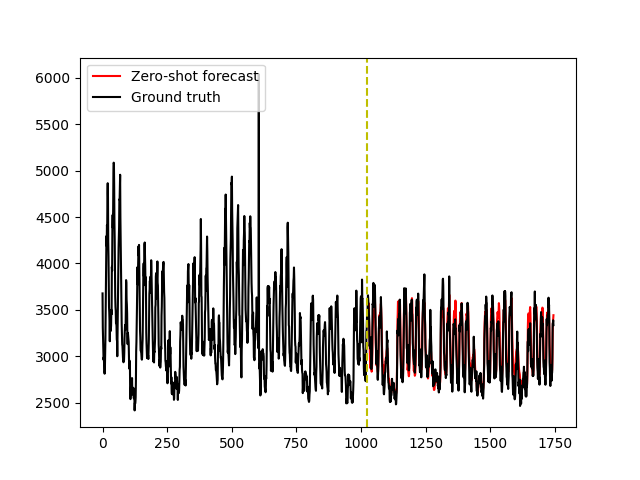}
  \caption{Electricity}
  \label{fig:sub3}
\end{subfigure}%
\begin{subfigure}{.37\textwidth}
  \centering
  \includegraphics[width=\linewidth]{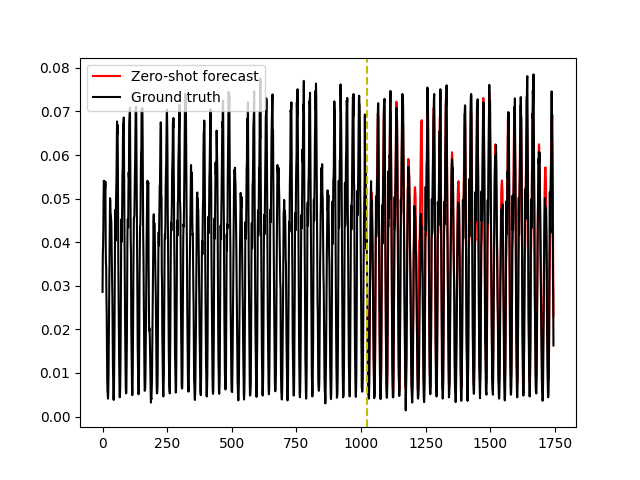}
  \caption{Traffic}
  \label{fig:sub3}
\end{subfigure}%
\\
\begin{subfigure}{.37\textwidth}
  \centering
  \includegraphics[width=\linewidth]{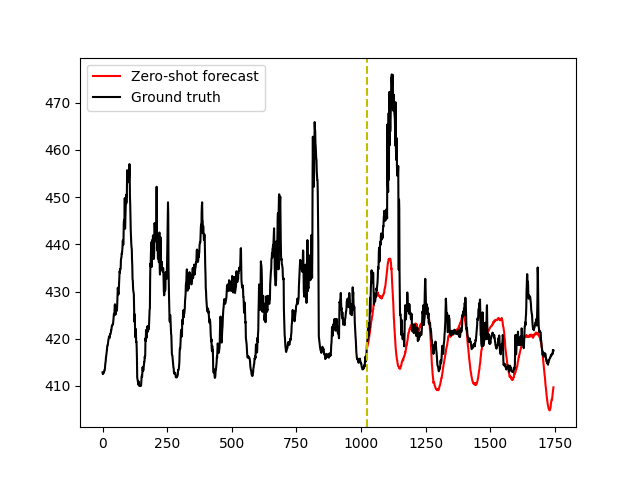}
  \caption{Weather}
  \label{fig:sub3}
\end{subfigure}%
\begin{subfigure}{.37\textwidth}
  \centering
  \includegraphics[width=\linewidth]{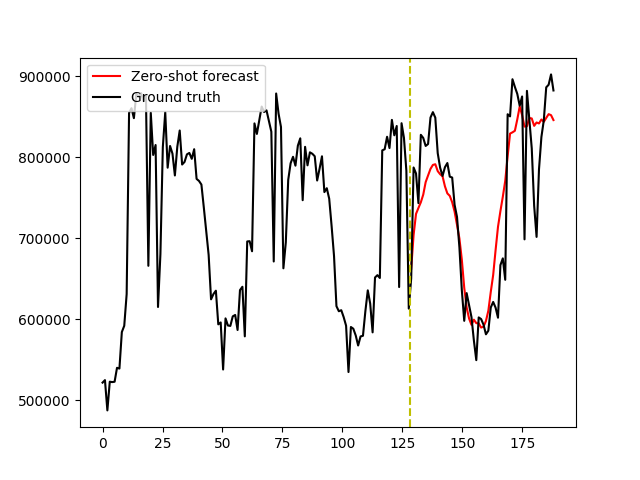}
  \caption{ILI}
  \label{fig:sub3}
\end{subfigure}%
\caption{Examples of GTT's zero-shot forecast results on benchmark datasets}
\label{fig:showcases}
\end{figure*}

To provide a clear view of GTT's zero-shot forecast performance, we give an example of GTT's zero-shot forecast results on each benchmark datasets in Figure~\ref{fig:showcases}. For ETT datasets, we show results on all variables. For other datasets, we only show results on the "OT" variable. 

% \begin{figure*}
% \centering
% \begin{subfigure}{.25\textwidth}
%   \centering
%   \includegraphics[width=\linewidth]{Figure_19.png}
%   \caption{input-1024-output-96}
%   \label{fig:sub3}
% \end{subfigure}%
% \begin{subfigure}{.25\textwidth}
%   \centering
%   \includegraphics[width=\linewidth]{Figure_20.png}
%   \caption{input-1024-output-192}
%   \label{fig:sub3}
% \end{subfigure}%
% \begin{subfigure}{.25\textwidth}
%   \centering
%   \includegraphics[width=\linewidth]{Figure_21.png}
%   \caption{input-1024-output-336}
%   \label{fig:sub3}
% \end{subfigure}%
% \begin{subfigure}{.25\textwidth}
%   \centering
%   \includegraphics[width=\linewidth]{Figure_22.png}
%   \caption{input-1024-output-720}
%   \label{fig:sub4}
% \end{subfigure}
% \caption{Prediction cases from the Traffic dataset}
% \label{fig:fig5}
% \end{figure*}

%%%%%%%%%%%%%%%%%%%%%%%%%%%%%%%%%%%%%%%%%%%%%%%%%%%%%%%%%%%%%%%%%%%%%%%%%%%%%%%

\end{document}